\documentclass{article}
\usepackage{times,epsfig,amssymb}
\usepackage{epstopdf,bm}
\usepackage{amsfonts}
\usepackage{subfigure}
\usepackage{spconf,amsmath,graphicx}
\usepackage{multirow}
\usepackage{color}

\title{Scalable Facial Image COMPRESSION with Deep Feature Reconstruction}

\name{Shurun Wang$^1$,   Shiqi Wang$^2$, Xinfeng Zhang$^2$, Shanshe Wang$^1$, Siwei Ma$^1$, Wen Gao$^1$}
\address{$^1$ Institute of Digital Media, Peking University, Beijing, China \\
$^2$Department of Computer Science, City University of Hong Kong, Hong Kong, China\\
}

\begin{document}

\maketitle

\begin{abstract}
In this paper, we propose a scalable image compression scheme, including the base layer for feature representation and enhancement layer for texture representation. More specifically, the base layer is designed as the deep learning feature for analysis purpose, and it can also be converted to the fine structure with deep feature reconstruction. The enhancement layer, which serves to compress the residuals between the input image and the signals generated from the base layer, aims to faithfully reconstruct the input texture. The proposed scheme can feasibly inherit the advantages of both compress-then-analyze and analyze-then-compress schemes in surveillance applications. The performance of this framework is validated with facial images, and the conducted experiments provide useful evidences to show that the proposed framework can achieve better rate-accuracy and rate-distortion performance over conventional image compression schemes.
\end{abstract}

\begin{figure*}[htbp]
\centerline{\includegraphics[width = 7.0in]{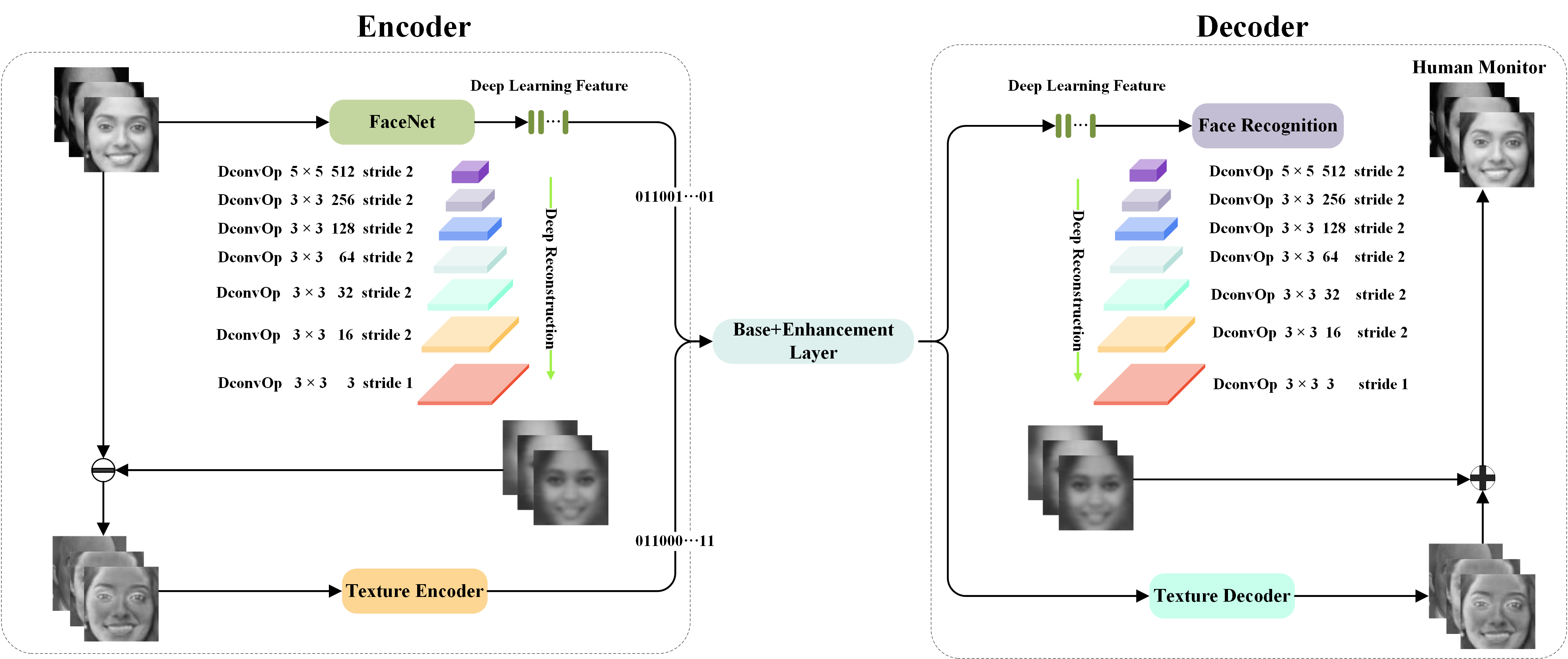}}
\caption{Illustration of the architecture of the scalable coding scheme. Every DconvOp in the framework is formulated with a de-convolutional, batch-normalization and activation layer. The kernel size, channel number and stride of every layer are listed alongside.}
\label{framework}
\end{figure*}

\section{Introduction}
\label{sec:intro}
There has been an explosive growth of image/video data and the related applications in recent years. With the development of smart city and intelligent security, surveillance data are of great importance as all kinds of activities can be recorded and analyzed in real-time.
Generally speaking, the surveillance camera data from the front-end are compressed and transmitted to the back-end for further analytical/understanding tasks, and this is referred to as the compress-then-analyze (CTA) paradigm.
In order to achieve higher compression ratio under the limitation of bandwidth, the low bit-rate compression is usually employed. However, the performance of analysis tasks may be significantly affected by the low bit-rate coding~\cite{gao2014ieee}. There is another feasible solution composed of feature extraction, compression and transmission, such that the compact features extracted at the front-end can be transmitted to the server side, and this is usually referred to as the analyze-then-compress (ATC) schema \cite{redondi2013compress}. 
Due to the features are much more compact compared to texture, the performance of the analysis task is quite promising at low bit-rate. 
However, it is difficult to recover the texture information as the features extracted from the original data cannot faithfully achieve signal level image/video reconstruction.

Feature coding algorithms play important roles in the ATC paradigm. In the literature, there have been many feature coding schemes proposed to further promote the compression ratio and transmission efficiency, based on both handcrafted features (e.g., SIFT~\cite{lowe2004distinctive}, SURF~\cite{bay2006surf, bay2008speeded}) and deep learning features.
A coding architecture for video sequence local feature compression was proposed by Baroffio \emph{et al.} in \cite{baroffio2014coding}.
Regarding the learning deep features, a trade-off between feature compression ratio and analysis performance was proposed in \cite{ding2017rate}.
The MPEG standards CDVS~\cite{duan2016overview} and CDVA~\cite{duan2018compact}, which standardize the handcrafted and deep learning features, have also found many applications in practice. Regarding surveillance video applications, besides the automatic visual analysis, human involved monitoring may also be required for further verification. As such, the texture reconstruction is also an important component which should not be ignored. From this perspective, Zhang \emph{et al.} proposed a framework for both feature and texture compression in \cite{zhang2017joint} and it provides the feasibility to compress the hand-craft features and video textures jointly. The joint rate-distortion optimization for simultaneous compression of texture and deep learning feature was further studied in \cite{li2018joint}. Moreover, the joint framework of texture-feature coding is overviewed and analyzed in~\cite{ma2018joint}, illustrating the advantages and disadvantages of both CTA and ATC schemes.

One aspect that has been largely ignored in feature and texture compression is that the information conveyed in feature is very helpful for image/video reconstruction. As such, to make better use of the deep learning features besides analysis tasks, we propose a scalable compression scheme based on the facial image, as face plays very important roles in video surveillance.
More specifically, the base layer conveys the deep learning feature, and a deep de-convolutional network is adopted for face reconstruction from deep learning features. The enhancement layer is responsible for the residuals between the input face image and reconstructed image from the feature, and the residuals are further compressed.
Extensive experiments have been conducted and the results show that the proposed schema can inherit the advantages of both ATC and CTA schemes.



\section{Scalable Compression Framework}
\label{sec:format}
In the traditional CTA paradigm, low bit-rate compression often results in severe distortions to the decoded texture information and the degradation of feature quality, which would lead to poor analysis performance. By contrast, for the ATC paradigm, it is difficult for human beings to view and monitor the image/video texture, which may limit its applications in real world. To address these issues, we propose a scalable framework, where the base layer is responsible for the feature and enhancement layer is responsible for the texture. There are three main advantages for this framework. First, when the texture reconstruction is not necessary, the base layer can be directly transmitted to support the ATC paradigm. As such, the fidelity of the deep feature can be guaranteed since the features are extracted from the original image.
Second, the redundancy between the feature and texture is exploited, which may significantly improve the texture compression performance. Third, it is feasible to perform the frequent analysis without image decoding and feature extraction, which is more economic in video surveillance applications.

The whole framework is shown in Fig.~\ref{framework}. The facial images are fed into the deep neural network for deep learning feature extraction, and the extracted features are compressed with direct quantization and entropy coding. Given the features of the facial images, our proposed scheme reconstructs the texture information with deep feature reconstruction, and the residuals between input and reconstructed texture are further compressed. In particular, the deep learning feature is a vector in a hyperspace with 128 dimensions.
Based on this consideration, we utilize de-convolutional neural network to recover the facial image as illustrate in \cite{mai2018reconstruction}. It is worth mentioning here that the DconvOP shown in Fig.~\ref{framework} is composed of a series procedures, including de-convolution layer, batch-normalization layer and ReLU activation (the activation function of the last layer is Tanh). After the deep reconstruction from the deep learning feature, we can acquire the residual information as the enhancement layer, which is further lossy compressed.
At the decoder side, the reconstructed image can be obtained by combining the deep feature reconstructed texture and the decoded residual from the enhancement layer. 

\section{Base and Enhancement Layer Compression}
\label{sec:pagestyle}
\subsection{Deep Feature Compression}
The deep feature extracted by FaceNet \cite{schroff2015facenet} is a vector in a hyper-space of 128 dimensions and every dimension is represented by a floating number. In order to compress and transmit the deep learning feature, as well as to guarantee the analysis performance, the deep feature undergoes quantization and entropy coding~\cite{mahoney2009data} for compression.
\subsection{Deep Feature Reconstruction}
\label{ssec:feature}
In \cite{mai2018reconstruction}, the authors proposed a deep feature reconstruction framework aiming to attack the face recognition system. We adopt this reconstruction strategy in this work.
The deep learning features are extracted by FaceNet \cite{schroff2015facenet}, which could embed the facial images into a point in the hyperspace. The distance among the coordinate points reflects the similarity of the images and we assume every dimension represents some certain characteristics of the human face. It is straightforward to adopt Mean Squared Error (MSE) as the loss but the information contained in the deep learning feature is mainly regarding the structure information, instead of the detailed texture. Therefore, we adopt the linear combination of the MAE (Mean Absolute Error) and the perceptual measurement as the loss function of the deep feature reconstruction network. The MAE can be formulated as

\begin{equation}
L_{MAE}(\bm{x}, \bm{x_{fea}})=\|{\bm{x}-\bm{x_{fea}}}\|_k=({\sum_{m=1}^{M}\|x_m - x_{fea,m}\|^{k}})^{\frac{1}{k}}.
\end{equation}
where $x$ and $x_{fea}$ represent the original facial image and the reconstructed image from the deep learning feature, respectively. Here, the parameter \(k\) is set to 1.

We adopt the feature map in VGG-19 model \cite{simonyan2014very} to compute the perceptual loss. The output of $\bm{ReLU3\_2}$ layer is utilized empirically to measure the structure information loss for the reconstructed image. The perceptual loss can be expressed as
\begin{equation}
L_{percept}(\bm{x},\bm{x_{fea}})=\frac{1}{2}\|F(\bm{x})-F(\bm{x_{fea}})\|_2^2
\end{equation}
As such, the loss function of the deep feature reconstruction network is given by,
\begin{equation}
Loss_{fea} = L_{MAE} + \lambda L_{percept},
\end{equation}
where $\lambda$ is the balancing factor between these two loss functions.

\subsection{Enhancement Layer Compression}
\label{ssec:texture}

For the enhancement layer, the compression of the residual $x_{resi}$ between the original facial image and the reconstructed image from the deep learning feature could be realized by traditional JPEG and JPEG2000 codecs. Moreover, the distribution of the residual patch could be different from common natural images and a deep learning based model can be utilized for the specific compression. As such, we also adopt the end-to-end compression framework in \cite{balle2016end}, which is based on generalized divisive normalization (GDN). The feature number of every convolutional layer for RGB images has been reduced from 192 to 128 here since compared to original images less structural information is contained in the enhancement layer.

The min-max normalization is applied to $x_{resi}$ to reveal the texture information $x_{tex}$ for compression and the minimum and maximum would also be encoded and transmitted to the decoder side for recovery of the texture scale,
\begin{equation}
    x_{resi} = x - x_{fea}
\end{equation}
\begin{equation}
    x_{tex} = \frac{x_{resi}-x_{min}}{x_{max}-x_{min}}.
\end{equation}
The loss function is the MSE between texture $x_{tex}$ and the decoded one $x_{rec}$,
\begin{equation}
    Loss_{tex} = \|\bm{x_{tex}-\bm{x_{rec}}}\|_{2},
\end{equation}
As such, the enhancement layer is conveyed and it is transmitted when necessary to reconstruct the texture in high fidelity.

\subsection{Implementation}
\label{ssec:Scheme Implement}

We implement the deep feature reconstruction model using TensorFlow. We initialize the network using the method in \cite{he2015delving} and the batch size is set to be 64. The learning rate is decayed exponentially from 0.01 to 0.0001 in 50 epoches and $\lambda$ is set to $0.00001$ to ensure the reconstruction of faces.

We can build our own residual dataset on the basis of the deep feature reconstruction network to train the end-to-end enhancement layer compression model. The optimization algorithm is the Adaptive Moment Estimation (Adam)~\cite{kingma2014adam} optimizer which is the same as the feature reconstructed model. The batch size is set to be 16 and the learning rate is 0.0002 for 10 epoches.

\section{Experimental Results}
\label{sec:experiment result}

In this section, we conduct experiments for validations in terms of both rate-accuracy and rate-distortion performance of the proposed framework.
The unconstrained large scale face dataset VGG-Face2~\cite{Cao18} is used for training. This dataset comprises
of over 3.3 million face images from 9131 subjects and there are over 360 images for per subject on average. The proposed framework is tested on a popular face verification dataset, Labeled Faces in the Wild (LFW)~\cite{LFWTech}. 

\begin{figure}[t]
\centerline{\includegraphics[width = 3.3in]{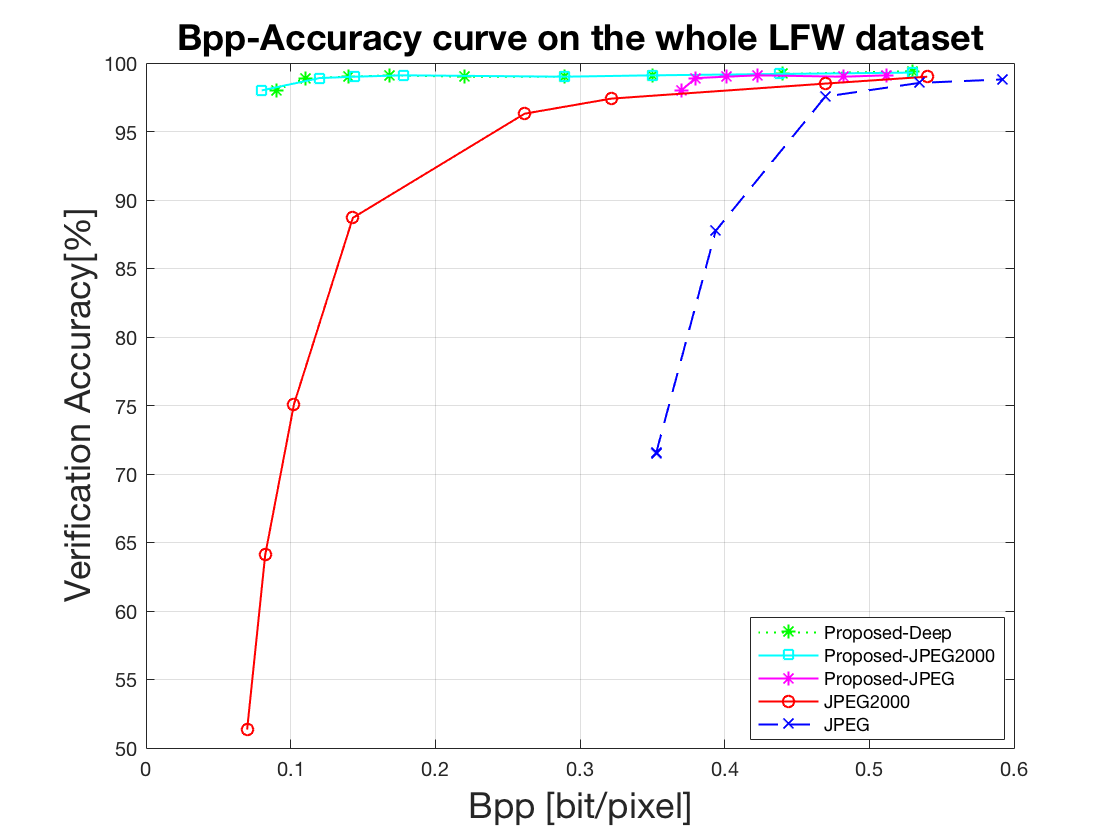}}
\caption{The rate-accuracy performance comparisons of the proposed scheme and other methods on the LFW dataset. The model for enhancement layer could be JPEG, JPEG2000 and deep learning based algorithms. Moreover, the number of coding bits here is obtained by the sum of the base layer and enhancement layer. }
\label{curve}
\end{figure}

\begin{figure}[htbp]
\centerline{\includegraphics[width = 3.4in]{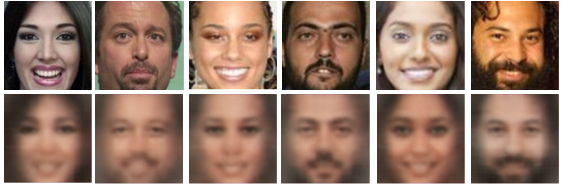}}
\caption{Examples of the deep learning feature reconstruction results from the base layer. First row: original images; second row: reconstructed images.}
\label{feature_image}
\end{figure}


\begin{figure}[htbp]

\begin{minipage}[b]{1.0\linewidth}
  \centering
  \centerline{\includegraphics[width=8.3cm]{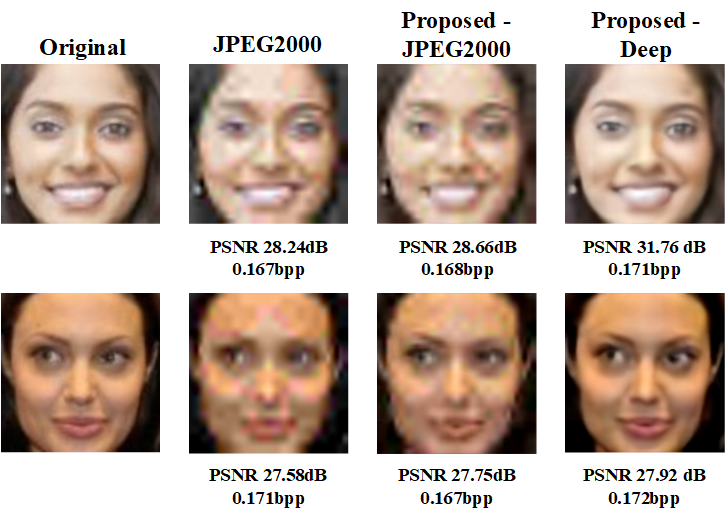}}
  \centerline{(a) JPEG2000}\medskip
\end{minipage}
\begin{minipage}[b]{1.0\linewidth}
  \centering
  \centerline{\includegraphics[width=8.3cm]{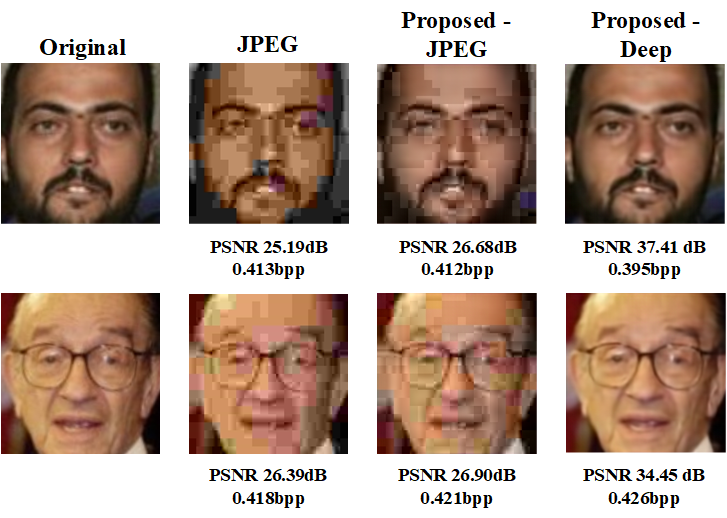}}
  \centerline{(b) JPEG}\medskip
\end{minipage}
\caption{Compression performance comparisons (anchor: JPEG/JPEG2000). The bit rate is obtained based on the sum of base and enhancement layers. For JPEG2000/JPEG, the enhancement layer is also compressed with the corresponding codec for a fair comparison.}
\label{rdper}
\end{figure}

\begin{figure}[htbp]
\begin{minipage}[b]{.48\linewidth}
  \centering
  \centerline{\includegraphics[width=4.2cm]{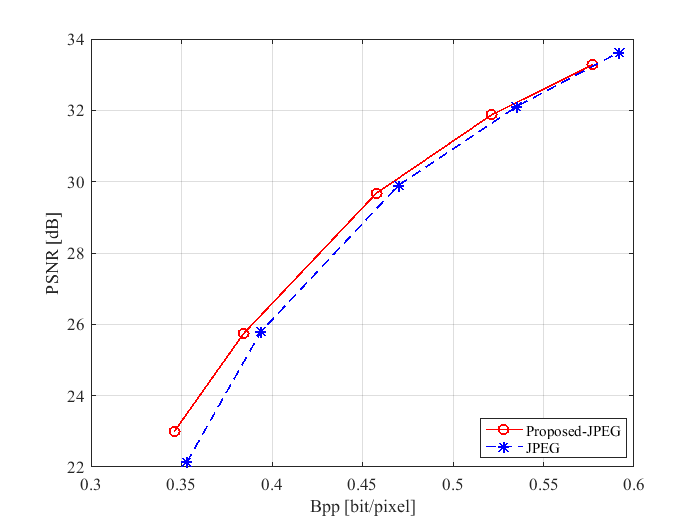}}
  \centerline{(a) JPEG}\medskip
\end{minipage}
\hfill
\begin{minipage}[b]{0.48\linewidth}
  \centering
  \centerline{\includegraphics[width=4.2cm]{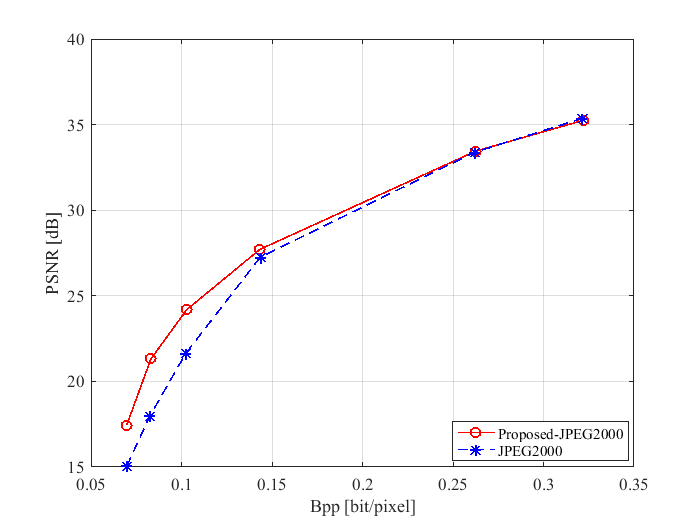}}
  \centerline{(b) JPEG2000}\medskip
\end{minipage}
\caption{Rate-distortion performance comparisons. (a) The enhancement layer is compressed with the JPEG; (b) The enhancement layer is compressed with JPEG2000. The statistics are obtained on the whole LFW dataset.}
\label{rd-curve}
\end{figure}

First, we evaluated the rate-accuracy performance of the proposed scheme. We compare the performance of the proposed framework with the CTA scheme that encodes the image with JPEG and JPEG2000. In particular, we adopt the deep learning based approach, JPEG and JPEG2000 to compress the enhancement layer. The experimental results are shown in Fig.~\ref{curve}.
It is obvious that the proposed scheme achieves the best performance. This is not surprising as it inherits the properties of the ATC approach, such that the features are extracted from the original texture. However, along with the degradation of the texture quality, the performance of the CTA schemes is significantly influenced, leading to lower analysis accuracy.

To gain more insights about the texture compression performance of our proposed scheme, we conduct experiments to evaluate the rate-distortion performance. The reconstructed images from deep learning features are shown in Fig.~\ref{feature_image}, which show that the main structure of the facial images have been preserved. Furthermore, we compare the compression performance and the subjective quality in Fig.~\ref{rdper}. For a fair comparison, when JPEG2000 is used as the anchor, we show the results when the enhancement layer is compressed by both JPEG2000 and deep learning framework. Again, for JPEG based comparisons, the enhancement layer is compressed by JPEG instead of JPEG2000. It is obvious that our proposed model improves the coding performance. Moreover, the rate-distortion curves shown in Fig.~\ref{rd-curve} provide useful evidence regarding the performance improvement in terms of PSNR.


\section{CONCLUSION}
\label{sec:conclusion}
In this work, we propose a scalable scheme for the facial image compression. The proposed scheme is composed of a base layer for feature compression and an enhancement layer for texture reconstruction, aiming at assimilating the advantages of both CTA and ATC. Interestingly, we find that the proposed scheme can inherit the rate-accuracy performance of ATC, and significantly improve the coding performance compared to traditional coding schemes. In the future, we will extend this framework to more domains in surveillance applications (e.g., vehicle and pedestrian).

\bibliographystyle{IEEEbib}
\bibliography{Template}

\end{document}